\documentclass[journal,singlecolumn]{IEEEtran}
\usepackage{times}
\usepackage{amssymb,bm,amsfonts,amsmath,mathrsfs}
\usepackage{graphicx}
\usepackage{subfigure}
\usepackage{caption}
\usepackage{url}

\usepackage{graphicx, epstopdf}
\usepackage{amssymb, amsmath}
\usepackage{commath}
\usepackage{setspace}
\usepackage{acronym}
\usepackage{mathrsfs}
\usepackage{ifthen}
\usepackage{textcomp}
\usepackage{float}
\usepackage{subfigure}
\usepackage[table]{xcolor}
\usepackage{color,soul}
\usepackage{cite}
\usepackage{setspace}
\usepackage{mathtools}
\usepackage{bm}
\usepackage{url}
\usepackage{xcolor}
\usepackage{empheq}
\usepackage{tikz}
\usepackage{subfigure}
\usepackage{stfloats,amsthm}
\DeclareMathSizes{12}{12}{7.7}{5.5}
\usepackage[bb=boondox,bbscaled=1]{mathalfa}

\begin{document}

\title{Model-Free Information Extraction in\\ Enriched Nonlinear Phase-Space}

\author{Bin~Li, Yueheng~Lan, Weisi~Guo, Chenglin~Zhao

\thanks{Bin Li and Chenglin Zhao are with the School of Information and Communication Engineering (SICE),
Beijing University of Posts and Telecommunications (BUPT), Beijing,
100876, China. Weisi Guo is with School of Engineering, University of Warwick,
West Midlands, CV47AL, and also with Alan Turing Institute, 96 Euston Road, London NW12DB, UK. Yueheng~Lan is with School of Science, BUPT, Beijing,
100876, China.}% <-this % stops a space
%\thanks{A. Nallanathan is with the Department of Informatics, King's
%College London, London, WC2R2LS, United Kingdom. (Email:
%nallanathan@ieee.org)}
%\thanks{Manuscript received Sept. 15, 2016.}
}

\maketitle

\begin{abstract}
Detecting anomalies and discovering driving signals is an essential component of scientific research and industrial practice.
Often the underlying mechanism is highly complex, involving hidden evolving nonlinear dynamics and noise contamination. When representative physical models and large labeled data sets are unavailable, as is the case with most real-world applications, model-dependent Bayesian approaches would yield misleading results, and most supervised learning machines would also fail to reliably resolve the intricately evolving systems.
Here, we propose an unsupervised machine-learning approach that operates in a well-constructed function space, whereby the evolving nonlinear dynamics are captured through a linear functional representation determined by the Koopman operator. This breakthrough leverages on the time-feature embedding and the ensuing reconstruction of a phase-space representation of the dynamics, thereby permitting the reliable identification of critical global signatures from the whole trajectory. This dramatically improves over commonly used static local features, which are vulnerable to unknown transitions or noise.
Thanks to its data-driven nature, our method excludes any prior models and training corpus. We benchmark the astonishing accuracy of our method on three diverse and challenging problems in: biology, medicine, and engineering. In all cases, it outperforms existing state-of-the-art methods. As a new unsupervised information processing paradigm, it is suitable for ubiquitous nonlinear dynamical systems or end-users with little expertise, which permits an unbiased excavation of underlying working principles or intrinsic correlations submerged in unlabeled data flows.
\end{abstract}


\IEEEpeerreviewmaketitle

\section*{Introduction}
Detecting anomaly and recovering driving signals is a ubiquitous general task when handling complex systems, with diverse applications in scientific research and engineering practice. According to the level of knowledge on system mechanisms, existing analysis tools are divided into two branches \cite{Bzdok2018Statistics}. If the mathematical description are completely available, the Bayesian approach, by far as the most primary analysis tool, would be used in data analysis and information extraction \cite{Bayes1763, neyman1942basic,winkler1972introduction}, e.g., inferring biological processes \cite{stahl2012bayesian,kharchenko2014bayesian,huelsenbeck2001bayesian}, interpreting natural wonders~\cite{Abbott2016Properties,Vitale2017Impact}, diagnosing patients~\cite{ashby2006bayesian,johnson1991bayesian,bretthorst1990bayesian}, analysing socioeconomic time-series~\cite{geweke1989bayesian}, processing signals in engineering \cite{Kay1993Fundamentals}. In contrast, if the complex mechanisms are partially perceptible or even impalpable, machine learning methods alternatively play to the advantage of distilling knowledge from unknown interacting structure \cite{Lecun1998Gradient,hinton2006fast}, which makes great successes in neuroscience \cite{Cadieu2014Deep}, computer vision \cite{lecun2015deep}, speech recognition \cite{hinton2012deep}, and named previously.

When applying Bayesian approaches, as Bayes himself put it \cite{Bayes1763}, one needs to know \emph{a priori} the model with more or less confidence, e.g. the relationship between hidden event and observation as well as the full associated statistics. It then incorporates prior knowledge into data analysis to infer the posterior beliefs \cite{Kay1993Fundamentals,pnevmatikakis2016simultaneous}, and supposedly to get the best results if a complete model is available under the ideal conditions of infinite past observations~\cite{neyman1942basic,Kay1993Fundamentals}. As George Box was keenly aware, however, ``all models are wrong, but some are useful" \cite{Box1976Science}. Since one may usually have no idea about what model might be suitable, the choice of models or the modeling precision will unavoidably affect the inferred results. Even for the simplest linear convolution system (see \textbf{1-e}), care should be exercised when computing the posterior beliefs of an unknown stimulus (see Fig. \textbf{1-d}). When the prior response is inaccurate (see Fig. \textbf{1-c}), for example due to its elusive expression or unknown parameters \cite{deneux2016accurate}, Bayesian inference (see Fig. \textbf{1-b}) tends to be risky, in particular when observational data is contaminated by noise and thereby mildly informative \cite{Elise2011Risk}, see Fig. \textbf{1-f}. As such, it might easily lead to discrepant results (see Fig. \textbf{1-a}, \textbf{1-d} and SI Fig. E1), and possibly misinterpreted mechanisms \cite{Elise2011Risk,pnevmatikakis2016simultaneous}.

More likely, there even exists no appropriate mathematical models for a large class of real problems, due to their high complexity \cite{deboer1987hemodynamic,Chuang1999Model,Ghasemloonia2015A,Kapitaniak2015Unveiling}. Often we are only left with rich observational data, rather than ubiquitous non-linear dynamics that govern system behaviors. Various learning machines, of course, provide a way of characterizing the alignment structure from input to output via a black-box manner. This is achieved by well-designed learning rules \cite{Lecun1998Gradient,hinton2006fast}, usually assisted by supervised training and consummate initialization~\cite{ haykin2004comprehensive}. Thus, they are able to learn local features from subset data and, if provided sufficient corpus, finally form a synergy to recognize specific patterns or understand intricate correlations. Unfortunately, the wider use of machine learning methods is largely hampered by its unfortunate property that, when a large amount of training corpus is unavailable or the alignment structure rapidly evolves, they would become unreliable as the test data involves unfamiliar structures excluded by previously incomplete training.

Thus, for many interesting and challenging problems, e.g. information processing in biological or engineering conditions, there is no generally accepted analysis tool to autonomously track and understand the intricately evolving dynamics buried in noisy data, whereby the underlying mechanism is enigmatic whilst the data-starving training is usually impractical.

\begin{figure*}[!t]
\centering
\includegraphics[width=13cm]{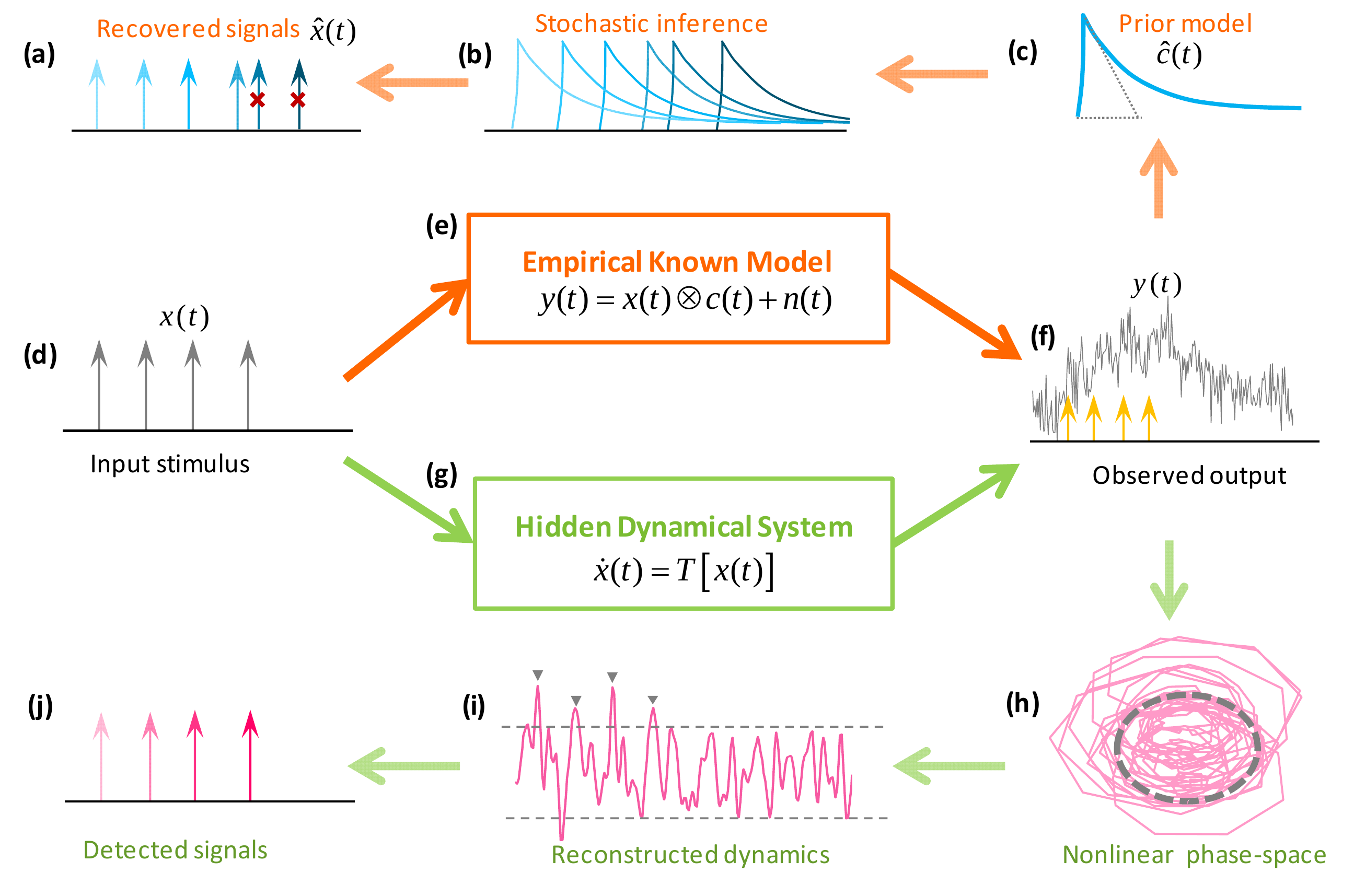}
\caption{\textbf{Model-specific Bayesian inference vs model-free information extraction} Noisy observations \textbf{(f)} were produced after input stimulus \textbf{(d)} passed through a system. In applying Bayesian statistical methods, an empirical model \textbf{(e)} was assumed to be known ($\otimes$ denotes the convolution operation). Based on the model response \textbf{(c)}, which usually requires much knowledge in specific scientific contexts, statistical inference was then implemented \textbf{(b)}, for example, by sequentially maximizing the posterior beliefs. When the prior model was incomplete, i.e. using $\hat{c}(t)$ other than $c(t)$ (with inaccurate parameters, see SI and Fig. E1-b), the eventually recovered signal would have errors \textbf{(a)} (i.e. two false inputs), potentially undermining the acquisition of the underlying mechanism. In contrast to model-dependent approach, our data-driven method assumed no prior model but the existence of certain unknown dynamics \textbf{(g)}. It is focused on a nonlinear attracting manifold built from the observational data \textbf{(h)}, followed by a low-rank reconstruction of the underlying dynamics \textbf{(i)} in a properly chosen observable space. Finally, it could detect target signals without prior expert knowledge \textbf{(j)}, and is thereby devoid of the bias risky, being valid even when the observation data was only mildly informative. }
\label{fig:2}
\end{figure*}
Here, we formulate a new perspective on unsupervised machine learning inspired by a data-driven concept, targeting at inferring crucial information from noisy response produced by complex non-stationary dynamics. We focus on reconstructing the nonlinear dynamics of observational data, and thereby discovering the global signatures of rapidly evolving systems via one-shot learning. To this end, we utilize a Koopman operator \cite{koopman1932dynamical} to transform the underlying nonlinear dynamics to an infinite-dimensional linear one. With the properly chosen time-feature embedding, a low-rank phase-space of the underlying nonlinear dynamics is then identified (see \textbf{1-h}). Finally, with a universally valid automatic determination of a threshold, the reconstructed trajectory is utilized to detect crucial signals (see \textbf{1-i} and \textbf{1-j}), i.e. via its low-dimensional local attracting or repelling skeleton sets on an approximately invariant manifold.

In this way, it excludes the requirement for governing dynamics or massive labels, and allows end-users with little expertise to extract information from complex responses. We test this unsupervised learning approach with real data from biology, medicine and engineering science, and show that it outperforms state-of-the-art methods. It creates a new paradigm for unsupervised information handling in real applications characterised by ubiquitous yet elusive nonlinear dynamics, which permits a supposedly unbiased understanding of system properties in disparate scientific contexts.

\section*{Results}

Our model-free approach directly exploits the nonlinear time-embedded dynamics of observational data, and implements autonomous detection of target information without any prior models or training. Starting with noisy observational data $y(t)$ (see Fig. \textbf{2-a}), it involves four steps, i.e., (1) constructing an enriched observable space via appropriate basis functions (see Fig. \textbf{2-b}), (2) obtaining a reduced representation of an orbit in phase-space (see Fig. \textbf{2-c}), (3) reconstructing the low-rank linear dynamics (see Fig. \textbf{2-d}), and (4) detecting information from the profile of an evolving trajectory (see Fig. \textbf{2-e} and \textbf{f}). In particular, (2) and (3) combines to forward the measurement of the current state to the next, thereby providing an alternative realization of the Koopman operator \cite{koopman1932dynamical}. The procedure is universally valid as demonstrated in the ensuing three models of different origins.

\subsubsection*{Multi-variate basis functions}
In order to build up a sufficiently rich observable space of the dynamics and construct an approximate invariant manifold, a set of multivariate time-series, namely basis functions or features, are generated from noisy data $y(n)$ (see \textbf{Fig. 2-a}, taking the neuron Ca$^{2+}$ fluorescence for example).

To be specific, we utilize the following inherent features of the noisy data (see \textbf{Fig. 2-b}), i.e. (1) a local convex shape, $y_2(n)$; (2) the mean difference between two neighbor regions (with distance $L$, see Method-B), $y_3(n)$; and (3) their energy ratio, $y_4(n)$. The original response $y(n)$ is used also as one feature, $y_1(n)=y(n)$. The designing of basis functions is critical to the information extraction, and here we give a general yet effective solution. See Method-B for its marked effectiveness in real applications.
\begin{figure*}[!t]
\centering
\includegraphics[width=18cm]{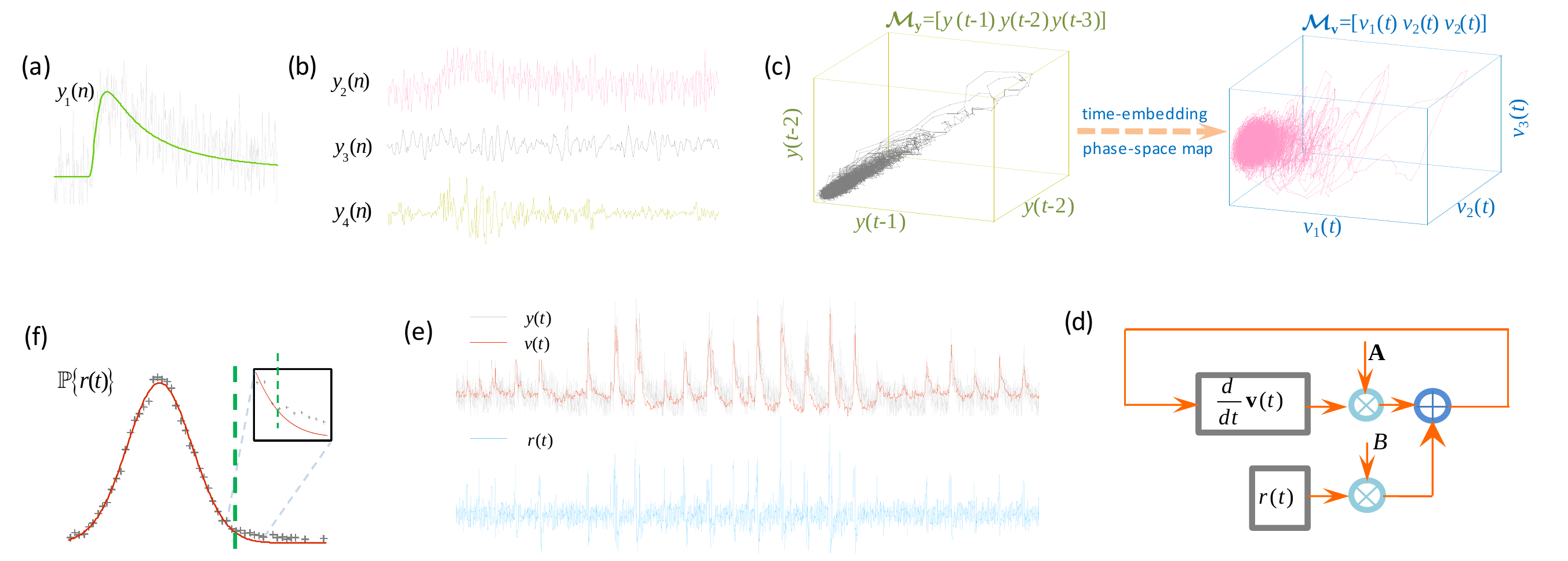}
\caption{ \textbf{Model-free information extraction in non-linear phase-space} The example is based on a measured Ca$^{2+}$ fluorescence of neuron cell. \textbf{(a)} Rather than an accurate transient shape, the prior model remains unknown, due to background noise and cell-specific response. \textbf{(b)} Multivariate basis functions were constructed from noisy measurement $y_1(n)$, including the local geometry convexity $y_2(n)$ ({\it top}), the mean shift $y_3(n)$ ({\it middle}) and the energy ratio $y_4(n)$ ({\it bottom}). \textbf{(c)} A manifold based on delay coordinates is mapped to another phase-space, by applying mode decomposition on the time-embedding Hankle matrix. A mapped phase-space $\mathcal{M}_v$ is represented by its three dominant modes, i.e., $[v_1(t),v_2(t),v_3(t)]^{T}$ \textbf{(d)} Linear resolvent analysis was used to identify modes that optimally describe the linear evolution trend. Rather than the random noise, we treat the $r$th component as a residual force, and the first $(r-1)$ components as one first-order Markov process. \textbf{(e)} For most stimuli/anomaly detection scenarios (note that, we are not intended to fully reconstruct the underlying nonlinear dynamics), $r$=2, i.e. $v(t)=v_1(t)$ and $r(t)=v_2(t)$. And, $v(t)$ provides a clean reconstruction of the noisy measurement $y(t)$. \textbf{(f)} A self-configured threshold for stimuli/anomaly detection can be automatically determined via the amplitude histogram of the decision signal.} %This threshold is set to the value whereby the histogram of amplitude began to deviate from a Gaussian density. }
\label{fig:2}
\end{figure*}

\subsubsection*{Trajectories on the approximately invariant manifold}
Provided the noisy data $y(n)$ ($n=nT_s$, $T_s$ is the sampling time), we treat it as the output of an hidden nonlinear system with dynamical state $\textbf{x}$, which is governed by:
\begin{align}
\dot{\textbf{x}}=T(\textbf{x}),
\end{align}
where $T(\cdot)$ denotes a nonlinear vector-valued function of the coordinate vector $\textbf{x}\in \mathbb{X} $ in a state space $\mathcal{M}$; $y=f(\textbf{x}):\mathbb{X}\rightarrow \mathbb{R}$ is the measured data. There may exist an attracting manifold built from the observable $y(t)$, e.g. represented by the delay coordinates $\mathcal{M}_y(t)\triangleq[y(t), y(t-T_s),y(t-2T_s)]^T$, as a low-dimensional (e.g. 3 dimensional) approximation to the original nonlinear dynamics, see \textbf{Fig. 2-(c)}. Then, our main purpose is to look for the best approximation, $\mathcal{M}_v(t)$, so that the omitted details play only minor roles. By applying the singular value decomposition (SVD) on a time-feature embedding Hankel matrix \cite{Broomheada1986Extracting} formulated by multivariate time-series $y_i(n)~(i=1,2,3,4)$ (see SI), $\textbf{Y}=\textbf{U} \bm{\Sigma} \textbf{V}^{*}$, a trajectory on this manifold is constructed via the first $r$ eigen-components of the right singular matrix $\textbf{V}$, i.e. $\mathcal{M}_v(t)\triangleq[v_1(t),v_2(t),v_3(t)]^T$.

\subsubsection*{Linear resolvent analysis}
Aiming to identify modes that optimally describe the overall evolution of observational data, a linearization is applied to the nonlinear dynamics. This process is known as linear resolvent analysis (LRA) in the context of nonlinear fluid dynamics \cite{mckeon2010a,sharma2016on}, or linear inverse model (LIM) in climate science \cite{penland1989random}, i.e.,
\begin{align}
\frac{d \textbf{v}(t)}{dt} = \textbf{A} \textbf{v}(t) + \textbf{B} r(t).
\end{align}

Here, $\textbf{A}$ and $\textbf{B}$ are the regression coefficients for the linear dynamics of an essential component $\textbf{v}(t)$ and a residual force $r(t)$. The residual force $r(t)$ can be chosen as either Gaussian noise \cite{penland1989random}, or one of the eigen-components \cite{brunton2017chaos}. For example, S. Burton et.al. used a similar first-order Markov model to split nonlinear chaotic dynamics into a linear patch and an intermittent driving \cite{brunton2017chaos}.

Recently, J. H. Tu et.al. showed that time-embedding mode decomposition, when combined with LRA/LIM~\cite{tu2014on}, could approximate the Koopman operator $\mathcal{K}$ \cite{koopman1932dynamical,mezic2013analysis,mezic2004comparison}. By forwarding the observation in the current time frame to the next, i.e. $\mathcal{K} T(\textbf{x}_n)=T(\textbf{x}_{n+1})$, the Koopman operator provides a way to identify the linearization transformation of complex nonlinear dynamics, implemented numerically via Koopman mode decomposition (KMD) or dynamic mode decomposition (DMD) \cite{rowley2009spectral,schmid2010dynamic}. Focusing on the time-evolution of observables of a dynamical system, it greatly facilitates information processing in the absence of a proper analytic description.

When it comes to reliable information extraction from data (rather than perfect reconstruction of the underlying dynamics), we find that, in most cases, the first component $v_1(t)$ is dominant \cite{Antoulas2002On}, e.g. the first eigen-value $\sigma_1$ is much larger than the remaining ones (e.g. $>$10-fold, see Fig. \textbf{3-a} and SI Fig. E3). After the time-delay state-space reconstruction, i.e. decomposition followed by linearization, the approximated dynamical component $v(t)$ provides a clean version of $y(t)$ \cite{Allen1997Optimal}, see Fig. \textbf{2-e} top, whereby the non-stationary baseline is removed (for the removal protocol, see SI Fig. E4) and ambient noise is suppressed, which enables further fine-grained analysis on temporal feature or system dynamics.

Other than the dominant component, we are especially interested in the $r$th component $v_r(t)$, which could be modeled as the external driving, i.e. $r(t)=v_r(t)$ \cite{ResidualForce} (the order $r$ may be automatically determined, see Method-D). In a mapped phase-space, the residual force $r(t)$ manifests the stimulus-induced transitions, which corresponds to unknown events or jumps to new patterns. Thus, a decision signal is defined as $d(t)=v(t)+r(t)$.

\subsubsection*{Automatic determination of the transition threshold}
In the absence of a stimulus, the driving signal is characterised by small fast-changing noise, and trajectories on the invariant manifold are limited to an attracting region, see Fig. \textbf{2-c}. When fed with a rarely occurring stimuli, a noticeable transition could be seen in $r(t)$, i.e., the trajectory makes an transition from the attracting region to another region, indicating a qualitative change of the linearized dynamics, see Fig. \textbf{2-c}, \textbf{3-e} \emph{bottom}, \textbf{4-c} and \textbf{5-c}. %As such, a decision signal follows a non-Gaussian distribution (see Fig. \textbf{2-f}).

Interestingly, in the context of information extraction, the amplitude histogram (or distribution) of the decision signal is characterised by a heavy decaying profile. This evidences that, produced by highly dynamic and complex systems, stimuli or abnormal transients usually exhibits a significant growth and decay in a short time, leading to a very considerable range of values. Specifically, we find a superposition of a Gaussian distribution $\mathscr{N}(d;d_0,\sigma_n^2)$ and an exponential distribution $\mathscr{E}(d-d_{th};\lambda)$ well fits the empirical histogram (see Fig. \textbf{3-b}, Method-C and SI Fig. E5). We may choose the decision threshold to be $d_{th}$, which is determined by checking the breaking point of the amplitude distribution (or empirical histogram), see Fig. \textbf{3-b}. Finally, target signals are detected by comparing $d(t)$ with the threshold $d_{th}$.
\begin{figure}[!t]
\centering
\includegraphics[width=9cm]{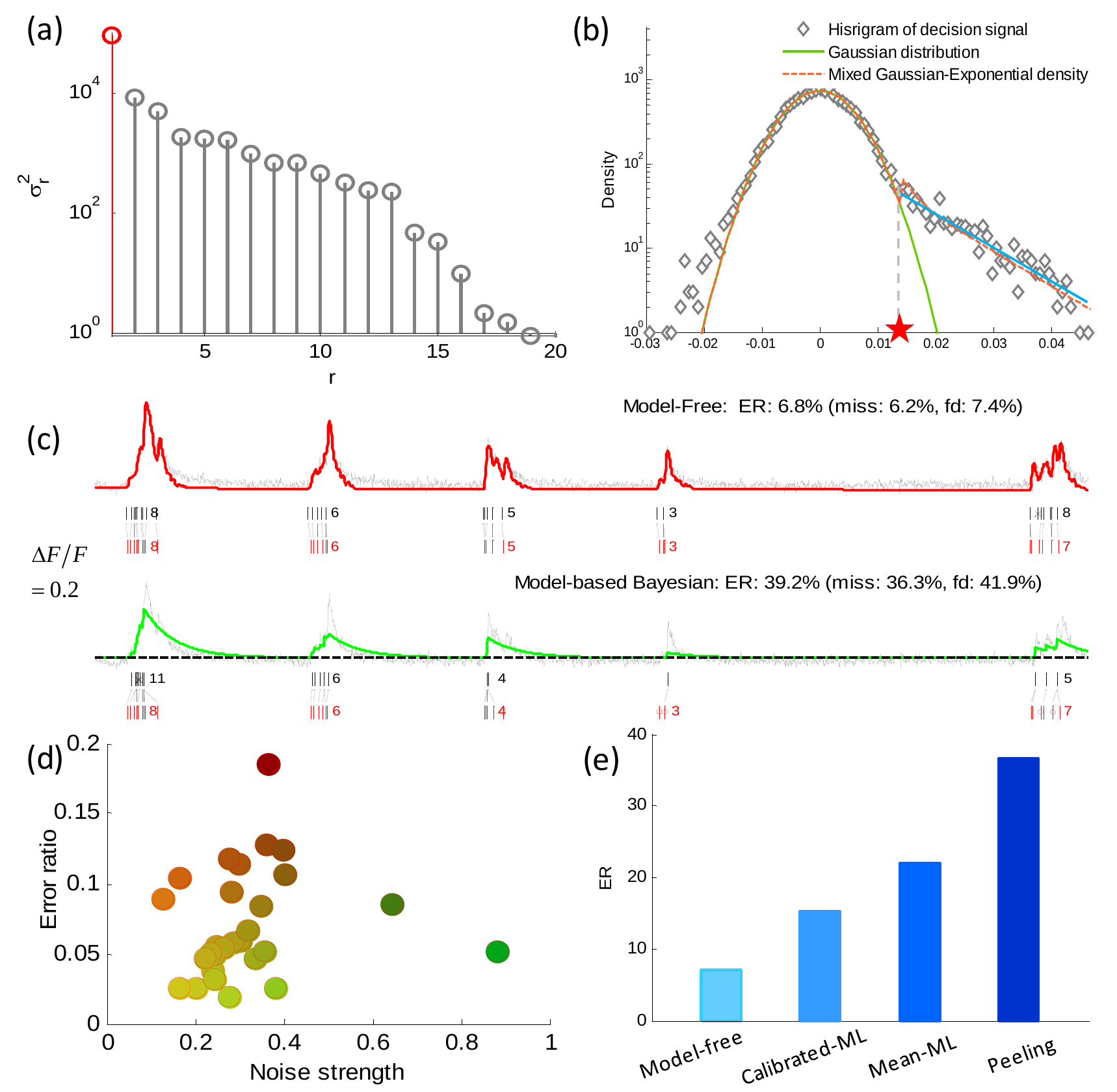}
\caption{ \textbf{Neuron spike detection from noisy Ca$^{2+}$ fluorescence with incomplete physiological model}. \textbf{(a)} Singular values of the time-feature embedding mode decomposition. The first singular value is dominant (measured neuron Ca$^{2+}$ fluorescence using the last-generation GECI, i.e. GCaMP6f, in vivo, $r=2$, $M=4$, $N=14400$, time duration 240 sec). \textbf{(b)} Amplitudes histogram of a decision signal. A mixture Gaussian-exponential distribution (scaled) conforms well with the empirical histogram. The decision threshold is then automatically determined (labeled with $\star$). \textbf{(c)} Spike detection results (root mean square noise variance 0.1, normalized by spike amplitude). \textbf{(d)} Averaged error ratio with various noise variances. For each measured spike train with different noise variances, the adaptive configuration was used and the noise strength needs not to be estimated. \textbf{(e)} Averaged error rate of our model-free approach in comparing with Peeling scheme and ML method whereby the key parameters (see SI Fig. E2) were extracted directly from the noisy fluorescence.  }
\label{fig:2}
\end{figure}
\subsection*{Case 1: Inference from an incomplete model}

We first consider the neuron spike detection from noisy Ca$^{2+}$ fluorescence, which is of great importance to recovery neuron firing stain and facilitate information processing in brain microcircuits. A well-known physiological model \cite{vogelstein2009spike,Brette2012Handbook} is widely adopted by current statistical Bayesian approaches, which involves Brownian motion induced transients, nonlinear overlapping/saturation, non-stationary baseline drift, ambient noise, and indicator-specific distortion (see SI Fig. E2). Besides a set of uncertain parameters, a general shape of individual spike remains also elusive (e.g. cell-specific or non-analytic), and an idealized exponential function with tuned rise time is often used \cite{deneux2016accurate,yaksi2006reconstruction}.

Real measurements were taken from a public repository (anaesthetized mouse, brain barrel cortex areas and in vivo), by using the last-generation genetically-encoded calcium indicators (GECIs), i.e. GCaMP6f~\cite{Akerboom2012Optimization,Chen2013Ultra}, see Material-A. Labelled spike stimulus was recorded via a cell-attached electrode. We verify our model-free approach on the measured responses ($M=4$, $r$=2), which is robust to nonlinear dynamics or unpredictable saturation effect (see SI and Fig. E6 for details). An averaged error ratio (ER) is $<$\text{ER}$>$=7.1\%, and the maximum ER of 18.5\%, ER$<$12\% in 90\% of the cases, see Fig. \textbf{3-e}.
\begin{figure*}[!t]
\centering
\includegraphics[width=16cm]{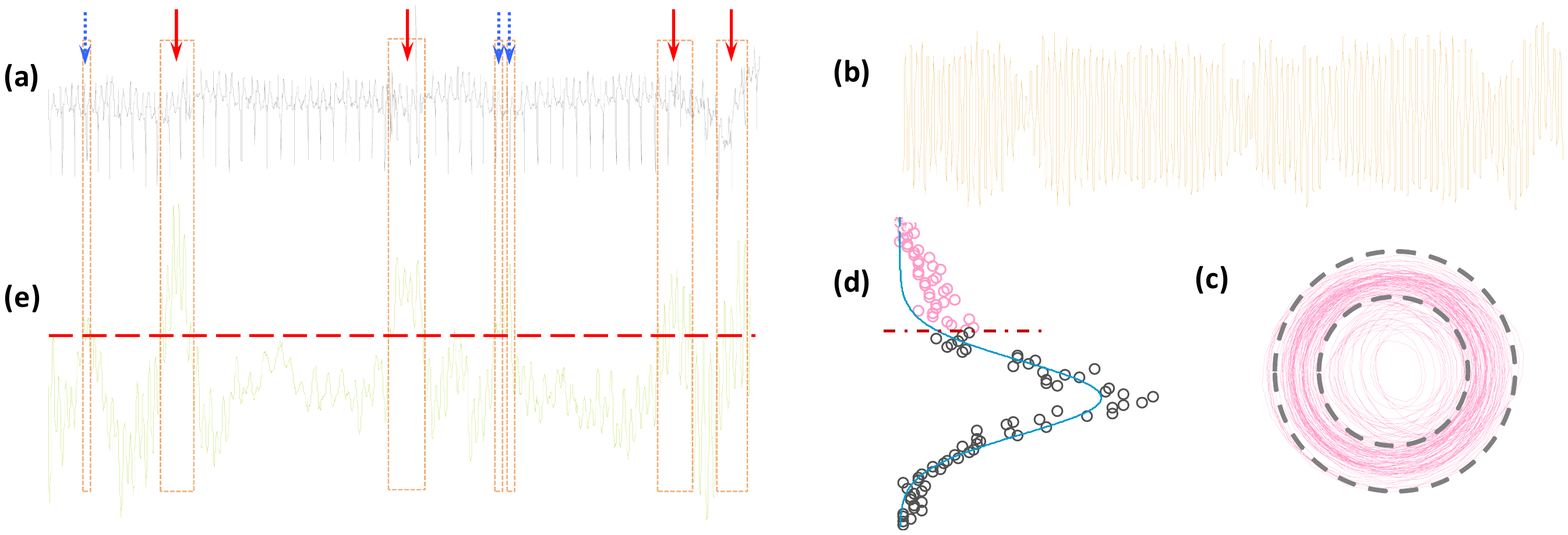}
\caption{ \textbf{ECG anomaly detection with unknown model and no labeled data} \textbf{(a)} Irregular ECG sequence without any label of age, sex and other information. \textbf{(b)} Linear component via a reconstructed low-dimension manifold ($r=2$, $M$=80). \textbf{(c)} A reconstructed low-rank phase-space, $\{v(t),r(t)\}$, consists of two regimes, i.e. the unstable inside circles driven by unknown force (corresponding to the irregular pattern of ECG) and the stable outside circles governed by the normal pattern. As a result, the extracted driven force \textbf{(e)} for irregular pattern can be extracted by performing a Hilbert-Huang transform on the decision signal $d(t)$. Again, a detection threshold can be automatically determined \textbf{(d)}, by checking the deviation between the empirical amplitude histogram and a fitted Gaussian distribution. }
\label{fig:3}
\end{figure*}

With the same data, the performance of another popular approach, i.e. ML detector, is evaluated, see Fig. \textbf{3-e}. Due to the incomplete modeling as well as the small amplitude of individual spikes, the acquisition of parameters (e.g. at least spike amplitude, decaying constant, baseline amplitude and noise variance, see Fig. E2) tends to be difficult for GCaMP6f. Using mean values of such parameters (i.e. the Mean-ML method), $<$\text{ER}$>$ is around 22.2\% ($<$20\% in 45\% of the cases). Further, another calibrated-ML method is tested, by extracting trail-specific parameters from noisy data \cite{deneux2016accurate}. With the normalised error 20$\thicksim$40\% in estimated parameters, then $<$\text{ER}$>$ is reduced to 15.1\%, with ER$<$20\% in 85\% of the cases. Another approach, Peeling method relying similarly on Bayesian inference \cite{Grewe2010High}, produces an averaged ER of 36.8\%. Other Monte-Carlo methods, e.g. sequential Monte-Carlo (SMC) \cite{vogelstein2009spike} and Markov-chain Monte-Carlo (MCMC) algorithms \cite{pnevmatikakis2016simultaneous}, fail to extract a unique spike train by only providing spiking probabilities or firing rate.

As the labeled spike stimulus are available, the supervised learning method, e.g. deep recurrent neural network (RNN), is also evaluated. We find the training accuracy of deep RNN is relatively high (by localizing multiple firing regions), and $\text{ER}_{\text{training}}$ is around 5.76\% (70\% data for training for one cell). Due to incomplete training and unknown structure in new data, the test accuracy would be seriously degraded and the resulting $\text{ER}_{\text{test}}$ is 22.9\%, while our unsupervised approach produces an $\text{ER}$ of 7.2\% for the same cell.

More importantly, with our model-free approach, ambient noise does not seem to be a key factor anymore limiting the performance, see Fig. \textbf{3-(d)}. This is strikingly different from Bayesian methods, whereby the noise level of data directly affects the accuracy of likelihood information \cite{Kay1993Fundamentals}, and seriously undermines the detection performance \cite{deneux2016accurate}. In this regards, our model-free approach may be profitably integrated into current analysing hardware, by effectively increasing their capability of data processing and breaking limitations of recording techniques (especially in noisy dynamical environments), which could greatly fuel the fast growing industry of biological inference, e.g. temporal spiking code of neurons or their piece-wise correlations \cite{Dettner2016Temporal}.

\subsection*{Case 2: Inference without a proper model}

Automatic anomaly detection of electrocardiography (ECG) wave is another challenging task. Despite the synthetic modeling efforts of ECG generator \cite{mcsharry2003a,deboer1987hemodynamic}, a complete stochastic model remains elusive, which needs to consider demographics, gender, age groups and separate mechanisms (e.g. ventricular fibrillation/flutter, or A-V block) for an accurate description. % complicating the task of ECG anomaly detection.

We challenge our model-free method with the recorded ECG wave \cite{wei2005assumption-free}, see Fig. \textbf{4-a}, $M$=70$\thicksim$90, $r=2$, sample length 2160. After reconstructing the nonlinear dynamics, we observe the representative phase-space, $\{v(t),r(t)\}$, may be divided into two zones (see Fig. \textbf{4-c}), i.e. the skirt ring and the central disk. In particular, the inside circle is excitable, which may be agitated by irregular changes, i.e. abnormal ECG patterns.

By performing a Hilbert transform on $r(t)$, a decision signal is designed as $d(t)=\mathsf{H}[v(t)+r(t)]$, see Fig. \textbf{4-e}. We immediately extract exciting patterns via an automatic threshold, see Fig. \textbf{4-d}. In comparing with Fig. \textbf{4-a} and \textbf{4-e}, the significant anomaly regions are accurately identified (for example, the 2nd, 3rd and 6th region, that were recognized by cardiologist). In contrast to other supervised learning methods, our approach is trail-specific and excludes also the data-starving training process, which hence avoids the unpleasant reliance on the vast amount of training data and the risk of bad generalization. As such, our unsupervised learning method is capable of operating on the data of small-sample size.

One popular unsupervised method, namely the chaos-game bitmap algorithm \cite{lin2003a}, is invoked for comparison, which, despite the claim of assumption-free, requires at least five tuned parameters \cite{wei2005assumption-free}. We find our new approach produces a sharper bound (see SI Fig. E8), which permits more accurate locating of anomaly patterns. Meanwhile, suspected irregular signatures ignored by the bitmap method are also identified (i.e. the 1st, 4th and 5th zone, see SI Fig E8), which involves recognizable distortion deserved for further medical investigation.

\begin{figure*}[!t]
\centering
\includegraphics[width=18cm]{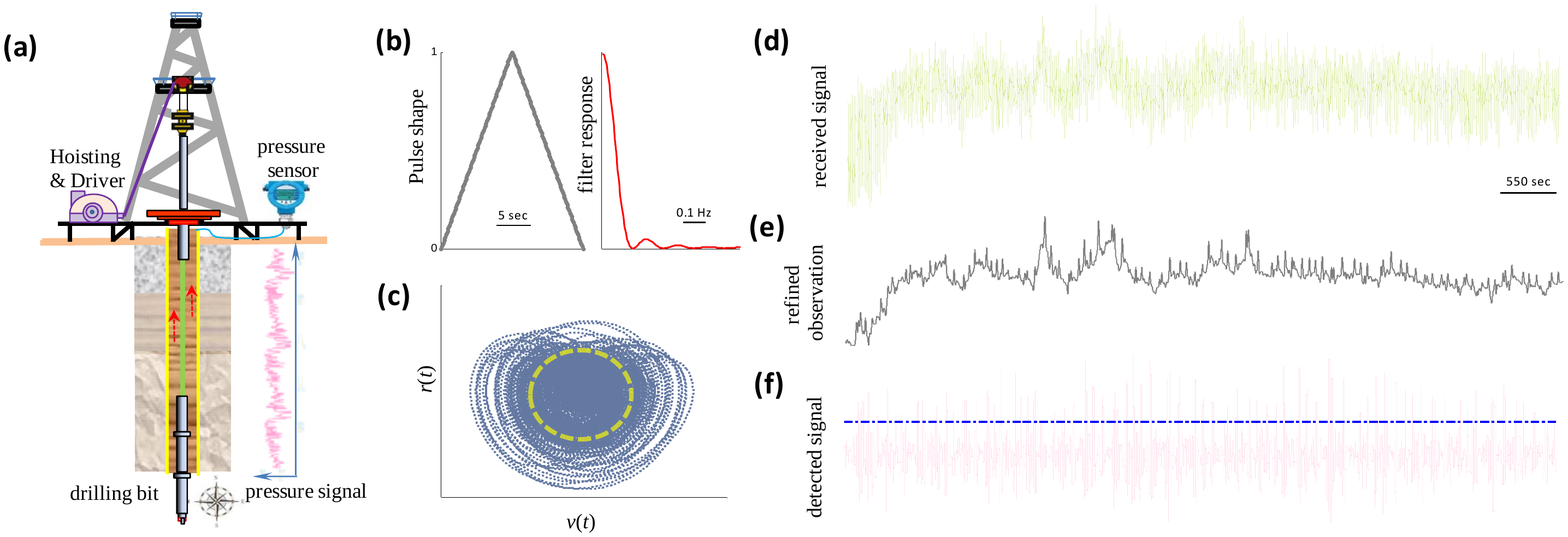}
\caption{ \textbf{Mud signal detection with unknown model and strong noise} \textbf{(a)} Schematic illustration of the MWD system. Real-time status information of underground drill-bit (including bit inclination, locality, advancing speed, etc.) was encoded via pulse intervals of mud pressure propagating to the ground, which was then recorded by a pressure sensor. \textbf{(b)} A rough pulse shape can be estimated, time-domain (\emph{left}) and frequency-domain (\emph{right}) response. \textbf{(c)} The reconstructed low-dimensional attractor. The trajectories escaping from the central part correspond to new stimuli, i.e. mud pulses. \textbf{(d)} Received mud signal. Besides strong noises aroused by mechanical vibrations as well as sensor error, the non-stationary random baseline drift further complicated signal detection (see SI Fig. E9). \textbf{(e)} Refined observations, i.e. the correlation output between noisy measurement and rough pulse in \textbf{(b)}. Despite the suppressed noise, random baseline drift still contaminates reliable signal detection. \textbf{(f)} Detected pulses from received noisy and fluctuated signal with our model-free approach ($r$=2, $M$=20, and see SI Fig. E10 for the self-configuration of threshold).  }
\label{fig:3}
\end{figure*}

\subsection*{Case 3: Inference with strong noise and no model}

Another difficulty with Bayesian approaches is the glitch produced by the prevailing noise, which, mixed with ubiquitous complex dynamics, seriously hampered reliable information processing. As a typical example in engineering, we study the detection of mud-pulse signals in measurement while drilling (MWD) systems (see Fig. \textbf{5-a}), whereby target information (e.g. bit inclination, locality, and other operating status of an underground drill) is mediated with the impulsive pressure of mud \cite{Brandon1999Adaptive}. Reliable recovery of such information is of great significance to informing control towards the goal of uninterrupted efficient drilling. %(drilling cease may cause even multimillion-dollar costs per day), as well as the adequate exploitation of natural resources (e.g. avoiding wellbore collision or side-wall collapse).

Contaminated by strong vibration noises \cite{Tucker1999An,Kapitaniak2015Unveiling} and nonlinear dynamics \cite{Ghasemloonia2015A}, serious distortion is inevitable in received signals, see the recorded measurement in Fig. \textbf{5-d}, drilling depth 5km, sampling time 100ms. Quite often dynamical features are case-specific for MWD systems (e.g. bit inclination, geological structure, etc), whilst general models remain unaccessible \cite{Brandon1999Adaptive}.
%Thus, one practical method was to use a frequency-domain filter. And, the resulting bit error rate (BER) was around 20.71\%.
Previous methods employ frequency-domain spectrum analysis to extract target signal \cite{Tu2012Research} (a rough pulse shape is available to guide the filter design, see Fig. \textbf{5-b} and SI Fig. E9), and thereby avail the dynamic control of drilling, aiming to ensure the successful exploitation of natural resources (e.g. avoiding wellbore collision or side-wall collapse). Unfortunately, the resulting bit error rate (BER) is as high as 20.71\%, due to unpredictable baseline drift and residual noise.

We test our model-free approach, and demonstrate again that it enables reliable pulse detection, even in the face of strong vibration noises and complex dynamics, with the BER of 6.94\% (see Fig. \textbf{5-f}, SI Fig. E9). More importantly, with the model-free approach, we are allowed to be relieved from building plausible yet unreal models and, alternatively, focus on the data-driven information processing with the aid of a reconstructed low-rank phase-space, which, as demonstrated, may be beyond the competence of model-dependent and training-based methods in real applications.

\section*{Discussions}
%Following the procedure of probing gently against the veil of the unknown, the boundaries of our human knowledge are constantly expanded.
Understanding new scientific principles always hinges on accurate information extraction and correct event interpretation.
One essential challenge across many disciplines is the lack of competent models or complete training for reliable information handling, which, more or less, hinders us from an unbiased recovery of the underlying mechanisms.
Despite significant advances of Bayesian-type or learning-based approaches, the reliance on \emph{a priori} models or training data limits their application and potentially underpins their accuracy, especially in biological, medical and engineering problems whereby accurate models and massive labels are usually difficult to produce.

Here, we develop a universal model-free information processing tool, which is able to extract crucial signals from noisy responses of a diverse set of challenging and important systems. As opposed to the model-specific and training-based methods, our new approach looks for a low-rank reconstruction of the invariant manifold, one that is determined by ubiquitous nonlinear dynamics but originally buried in harsh noises. Information extraction is thereby implemented by examining the broad evolving trends of trajectories.
As such, it opens up a new paradigm for data-driven inference and unsupervised learning, and allows us to be freed from building partially correct models or massive labels, which, more importantly, permits unbiased understanding of the underlying truth in natural or engineering world.

\subsection*{Material and Method}
\subsubsection*{\textbf{A.~~Database}}
The GCaMP6 data used in spike detection is available at http://crcns.org/. The ECG data used for anomaly detection are available at http://www.cs.ucr.edu/$\thicksim$wli/SSDBM05/. All other data are available from the authors upon request.

\subsubsection*{\textbf{B.~~Multi-variate Basis Function}}
In constructing multivariate basis functions, two neighboring sectors were considered. I.e., for a given time index $n$, the left sector was specified by $\mathbb{R}_0=[n-L/2, n-L/2+1,\cdots, n-1]$, while the right sector was $\mathbb{R}_1=[n+1, n+2,\cdots, n+L/2]$, where $L/2$ denotes the regime width (see Method section $D$ for its configuration). Provided the measured data sequence $y(n)~(n=0,1,\cdots,N-1)$, the fourth feature time-series was itself, $y_4(n)=y(n)$. (1) The first feature time-series, i.e. a local convex shape, was designed as $y_2(n)\triangleq y(n)-\frac{1}{L}  [ \sum_{n_0 \in\mathbb{R}_0} y(n_0) + \sum_{n_0 \in\mathbb{R}_1} y(n_0) ]$. (2) The second feature time-series, i.e. the mean difference, was designed as $y_3(n)\triangleq \sum_{n_0 \in\mathbb{R}_1} y(n_0)-\sum_{n_0 \in\mathbb{R}_0} y(n_0)$. (3) The third feature time-series, i.e. the energy ratio, was designed as $y_4(n)\triangleq \sum_{n_0 \in\mathbb{R}_1} y^2(n_0)\big/\sum_{n_0 \in\mathbb{R}_0} y^2(n_0)$.

In applications, slight changes can be incorporated when formulating basis functions. First, in the presence of Gaussian type noise, the above mean-value based feature time-series is adequate. For other unknown noises (e.g. impulsive noises as in MWD systems due to sudden strong mechanical vibration), the median-value based multivariate time-series may be similarly constructed, e.g. with the mean-value term $\sum_{n_0 \in\mathbb{R}_0} y(n_0)$ replaced by $\texttt{median}\{y(n_0),n_0 \in\mathbb{R}_0\}$. Second, the pretreatment of noisy observation $y(n)$ will be beneficial. When the roughly estimated pulse shape $w(n)$ is available (see Fig. \textbf{5-b}), as opposed to directly constructing features from original data $y(n)$, a refined observation was firstly derived by $\hat{y}(n)=y(n) \odot w(n)$ (where $\odot$ denotes the correlation operation, see Fig. \textbf{5-e}). Then, featured time-series will be derived from $\hat{y}(n)$.
%if some partial knowledge (maybe not accurate) could be available, it can be also casted into our multi-variate features. Taking the rough pulse shape $w(n)$ for example (see Fig. \textbf{5-b}), as opposed to directly constructing features from original data $y(n)$, a refined observation was firstly derived by $\hat{y}(n)=y(n) \odot w(n)$ (where $\odot$ denotes the correlation operation, see Fig. \textbf{5-e}). Then, featured time-series will be derived from $\hat{y}(n)$.

\subsubsection*{\textbf{C.~~Threshold Adaption}}

The histogram of a decision signal $d(t)$ can be determined directly. If normalised, this empirical histogram provides an estimation of the probability distribution function of decision signals, which was characterized by a heavy decaying profile. A well-fitted mixture Gaussian-exponential distribution is given by:
%\begin{align}
%p\{d(t)=d\}= w_G \cdot\frac{1}{\sqrt{2\pi\sigma_d^2}}\text{exp}[-(d-d_0)^2/(2\sigma_d^2)] + (1-w_G ) \cdot \lambda \text{exp}[-\lambda(d-d_{th})]
%\end{align}
\begin{align}
\text{Pr}\{d(t)=d\}= w_G \cdot \mathscr{N}(d;d_0,\sigma_d^2) + (1-w_G ) \cdot  \mathscr{E}(d-d_{th};\lambda),
\end{align}
where $w_G$ denotes the weight coefficient of a Gaussian distribution; $d_0$ and $\sigma_d^2$ for the mean and variance of the Gaussian distribution. $\lambda$ is the decay constant and $d_{th}$ gives the minimal value of a shifted exponential distribution (see Fig. E5). As observed, the threshold corresponds exactly to the non-Gaussian detachment point of an estimated probability distribution $\text{Pr}\{d(t)=d\}$, i.e. $d_{th}$.

The data-driven estimation of $d_{th}$ from $d(t)$ was thereby straightforward. First, using the symmetry property of a Gaussian distribution, we obtained a rough estimation, i.e. $\hat{d}_{th}$, which is chosen by minimizing a symmetry metric $\sum_{d<\hat{d}_{th}}(\text{Pr}\{d-d_0\} - \text{Pr}\{d+d_0\})/\text{Pr}\{d+d_0\}$. Then, the exclusion region $\mathbb{R}_E=\{d(t)<-\hat{d}_{th},d(t)>\hat{d}_{th}\}$ are determined, and the fitted Gaussian distribution, $\hat{\mathscr{N}}(d;d_0,\sigma_d^2)$, was derived. Finally, the detection threshold was determined by checking the difference between a fitted Gaussian distribution and the empirical density (see Fig. \textbf{3-b}).

\subsubsection*{\textbf{D.~~Implementations}}
In applications, the involved parameters of our universal model-free method can be configured automatically. First, the length of neighboring sectors in constructing multivariate basis function (i.e. $L$) was supposed to be large in order to suppress noise. However, a too large length may smooth out the abrupt transitions. In practice, it is configured according to the minimal interval among two stimuli. Second, the memory length $M$ in formulating a time-embedding Hankel matrix (see SI) and the analysis order $r$ can be determined, for example, by minimizing mean square error between a decomposed component $\textbf{v}(n)$ and the reconstructed linear term $\hat{\textbf{v}}(n)$, i.e. $(M,r)=\arg\min  ||\hat{\textbf{v}}(n;M,r) - \textbf{v}(n;M,r)||_2^2$ (see SI Fig. E5).

\subsubsection*{\textbf{Acknowledgements}}
We thank Dr. Z. K. Wei for his excellent research assistance. C. L. Zhao was funded by the International Exchanges Scheme of National Natural Science Foundation of China (NSFC) and Royal Society (Grant 6151101238). Y. H. Lan was funded by National Natural Science Foundation of China (Grant 11375093).

\subsubsection*{Author contributions}
B. Li conceived the idea and sourced the data, B. Li, W. S. Guo, Y. H. Lan designed, constructed, and analysed the algorithm systems. B. Li, W. S. Guo, Y. H. Lan, and C. L. Zhao together interpreted the findings and wrote the paper.

\bibliography{scibib}

\begin{thebibliography}{10}
\bibitem{ResidualForce}
More generally, we may be interested with the following $R\geq r$ components,
  and we further formulated a generalized residual force $\tilde{r}(t)=
  \sum_r^R \sqrt{\sigma_i} v_i(t)/\sum_r^R \sqrt{\sigma_i}$ (see SI Fig. E6),
  where the residual order $R$ is related with the distribution of eigen-values
  as well as the reconstruction error of $y(t)$.
\end{thebibliography}

\bibliographystyle{IEEEtran}

\end{document}